%% file: root.tex
\documentclass[letterpaper, 10pt, conference]{ieeeconf}
\IEEEoverridecommandlockouts
  \usepackage{amsmath}
  \usepackage{amssymb}
  \usepackage{amsfonts}
  \usepackage{graphicx}
  \usepackage{color}
  \usepackage{subcaption}
  
  \newcommand*\diff{\mathop{}\!\mathrm{d}}
  
  \allowdisplaybreaks

  \input{macros}

  \pdflatex

  \newcommand{\pre}[1]{{\small\textsf{#1}}}

  \renewcommand{\path}{{\text{path}}}
\newcommand{\goal}{{\text{goal}}}
\newcommand{\switch}{{\text{switch}}}

\renewcommand{\baselinestretch}{.995}

\newcommand{\danny}[1]{\textcolor{blue}{#1}}
\newcommand{\JS}[1]{{\color{red}JS: {#1}}}
\renewcommand{\hide}[1]{}

\title{\LARGE \bf
Describing Physics \emph{For} Physical Reasoning:\\Force-based Sequential Manipulation Planning
}

\author{Marc Toussaint$^{1,2}$ and Jung-Su Ha$^{1,2}$ and Danny Driess$^{1,3}$%
  \thanks{$^{1}$ Max Planck Institute for Intelligent Systems, 70569 Stuttgart, Germany. $^{2}$ Intelligent Systems Lab, TU Berlin, 10587 Berlin, Germany. $^{3}$ Machine Learning \& Robotics Lab, University Stuttgart, 70569 Stuttgart, Germany.
}%
}

\begin{document}

\maketitle
\thispagestyle{empty}
\pagestyle{empty}

\begin{abstract}
  Physical reasoning is a core aspect of intelligence in animals and
  humans. A central question is what model should be used as a basis
  for reasoning. Existing work considered models ranging from
  intuitive physics and physical simulators to contact dynamics models
  used in robotic manipulation and locomotion. In this work we propose
  descriptions of physics which directly allow us to leverage
  optimization methods for physical reasoning and sequential
  manipulation planning. The proposed multi-physics formulation
  enables the solver to mix various levels of abstraction and
  simplifications for different objects and phases of the solution. As
  an essential ingredient, we propose a specific parameterization of
  wrench exchange between object surfaces in a path optimization
  framework, introducing the point-of-attack as decision variable. We
  demonstrate the approach on various robot manipulation planning
  problems, such as grasping a stick in order to push or lift another
  object to a target, shifting and grasping a book from a shelve, and
  throwing an object to bounce towards a target.
\end{abstract}


\section{Introduction}

Reasoning is an essential form of generalization in AI systems,
implying decision making competences in situations that are not part of
the training data. Understanding the structure of reasoning problems in
the real world therefore yields important insights in what kind of
structure we might want to impose on learning systems for strong
in-built generalization.

In this work we aim to contribute towards general-purpose physical
reasoning, by which we mean performing inference over unknowns or
controls given a model of physics, and constraints or objectives on
(future) configurations. We believe that a core question is \emph{how
  to model physics} for the purpose of physical reasoning. In other
words, perhaps the pressing challenge is not to develop algorithmic
solvers for \emph{any} kind of forward model $\ddot x=f(x,\dot x,u)$
of physics, or any kind of physical simulator. Instead, the challenge
is to formulate specific models and abstractions of physics that are
appropriate for physical inference. We particularly touch on the
issues of multi-physics descriptions and exposing a logic
of physical interaction.

\emph{What are appropriate models of physics for enabling physical
  reasoning?} Physics is usually described as a differential equation
$\ddot x=f(x,\dot x,u)$ and simulation as numerical forward
integration. In such forward descriptions of physics, reasoning
becomes a problem of \emph{inverting physics}, inferring decisions
that lead to desired future configurations.
Physics can also be described via constraints on correct paths, which
we refer to as a \emph{path description of physics}. Of course, any
differential equation description of physics directly implies also a
path description, namely by imposing the equality constraint $f(x,\dot
x,u) - \ddot x= 0$ on the path. And a path description can, by
iteratively solving only short horizon problems (MPC-like), be used to
implement a forward simulator of physics.

In the case of contacts, forward models often describe $f$
itself as a mathematical program, i.e.\ via a linear complementary
problem
\cite{1997-Anitescu-Formulatingdynamicmultirigidbody,2017-Fazeli-Parametercontactforce},
or a Gaussian principle \cite{2012-Todorov-MuJoCophysicsengine}. This
is no issue for forward integration in simulators, and can also be
made (piece-wise!) differentiable based on classical sensitivity
analysis of NLP solutions
\cite{1985-Fiacco-Sensitivityanalysisnonlinear,1995-Levy-SensitivitySolutionsNonlinear,2010-Izmailov-SolutionsensitivityKarush}. But
when translating this to path constraints this implies an equality
constraint that itself contains a local mathematical program, which is
known as bi-level optimization and makes long-term physical reasoning
hard. Posa \cite{2013-Posa-Direct} thoroughly discussed the benefits
of direct path optimization over bi-level optimization, with which we fully agree.


\emph{Is there just one correct model of physics for reasoning?}  The
sciences describe physics on many levels of simplification: quantum
field dynamics, fluid dynamics, rigid body Newtonian physics,
quasi-static physics, toy-like physics as in some games, and intuitive
conceptions of physics we find in humans
\cite{2018-Smith-Differentphysicalintuitions}. It would be too
limiting for physical reasoning to make use of only one particular
abstraction of physics, or one particular physical simulator. Complex
reasoning requires the reasoning process to deliberately apply
different levels of simplification for different aspects of
inference. One approach to enable this is by describing physics itself
as if it would switch laws, as if physics would decide itself that
certain objects follow sometimes Newtonian laws, while others follow
blocks-world pick-and-place laws, and yet others follow quasi-static
equations. We use the term \emph{multi-physics}\footnote{In the
  numerical simulation sciences, the term is used for simulations that
  involve multiple physical models, or multiple simultaneous physical
  phenomena, or multiple components where each is governed by its own
  principle(s). This typically refers to fully different types of
  physical interactions, such as fluid dynamics, electromagnetism, or
  chemical interactions. Our use of the term is yet more limited to
  just mixing stable and dynamic modes, rather than completely
  different areas of physics. But in both cases it conceptually means
  to leverage multiple physical models for physical modeling and
  reasoning.} to refer to this approach.



\noindent The contributions of this work are as follows:

1) We propose a novel approach to introduce decision variables for the
wrench exchange between object surfaces in a path optimization
framework. Our parameterization of wrenches, based on the
\emph{point-of-attack} (POA), can naturally handle any contact geometries
(e.g., also line-to-line or point-to-surface geometries) and
continuous transitions between them, and thereby allows the optimizer
to find wrench interactions consistent with both, contact geometry and
the Newton-Euler equations.

2) We propose a multi-physics approach to allow the optimizer to use various models of physical interaction for
general physical reasoning, which include and integrate general
force based interactions, quasi-static dynamics, and
pick-and-place type interaction modes.

3) We extend Logic-Geometric Programming \cite{18-toussaint-RSS} to
incorporate these interaction models and introduce additional decision
variables for time-scaling between discrete configurations. In this
way we enable joint optimization the path of configurations and their
real-time scaling, which is essential as physics constraints can only
be fulfilled when co-optimizing the timing of physical interactions.

We integrate these methods in a path optimization framework that takes
a skeleton (a logic specification of the sequenced interactions) as
input and tries to solve for a physically feasible and optimal path
\cite{17-toussaint-Newton,15-toussaint-IJCAI,17-toussaint-ICRA,18-toussaint-RSS}.
Correct paths mix interaction modes of different abstraction levels
for different object pairs in different time intervals, and we
optimize full manipulation sequences across such switches in
description. Using this, we demonstrate the approach on sequential
dynamic and quasi-static manipulation problems, such as pushing with a
stick, lifting a ring with a stick, toppling over a box, and sliding a
book from a shelf before grasping. The breadth of tasks highlights the
generality of the formulation.

After discussing related work in Sec.~\ref{secRelatedWork}, we detail
our modeling approach in
Sec.~\ref{secModel}. Sec.~\ref{secExperiments} presents our
demonstrations.

\section{Related Work}\label{secRelatedWork}

\subsection{Trajectory optimization through contacts}

Trajectory optimization through contact interactions have previously
been considered for footstep planning, dynamic locomotion, and
manipulation
\cite{2014-Deits-Footstepplanninguneven,2013-Posa-Direct}. Specifically, in \cite{2013-Posa-Direct}
point-to-point contacts are considered (patches are represented as
multiple contact points) and one fundamental type of interaction
(complementary contacts) is modeled. In comparison, our multi-physics
formulation allows to mix interaction modes and our POA approach
parameterizes the wrench between any two object surfaces, naturally
including all contact geometries (point-to-point, point-to-line,
line-to-line, point-to-surface, line-to-surface, surface-to-surface)
and continuous transitions between them (see the box sliding example).

Contact-invariant optimization \cite{2012-Mordatch-Contactinvariant}
has demonstrated impressive sequential manipulation plans. It relies
on the pose difference between contact frames ($e_{i,t}(s)$ in their
notation), which means that the supposed contact points (and contact
geometry) have to be pre-fixed on the endeffector. Further, the
contact invariant objective does not allow for sliding (because the
relative pose velocity is penalized). Therefore it cannot not be
applied to sliding or using a tool to exert forces (as no dedicated contact
endeffector frame on the tool can be ad hoc fixed), which are both
inherent in our demonstrations.

The core of our approach is to introduce a general parameterization of
contact interactions that allows the optimizer to find wrench
interactions consistent with both, contact geometry and the
Newton-Euler equations.


\subsection{Wrench exchange parameterization}


A core question in describing force based interactions in path
optimization is how precisely decision variables are introduced to
represent wrench exchange. Fazeli et
al.~\cite{2017-Fazeli-Parametercontactforce} considered general
transmission of wrenches through contact patches or multiple
contacts. However, in a path optimization setting, the number of
contact points depends on the current geometry and is variable
throughout optimization. To address this, Xie et
al.~\cite{xie2016rigid} introduced the concept of the equivalent
contact point, which subsumes wrenches exchanged via a line or surface
contacts into a single contact point. With $\l$ the forces \hide{or
  wrench? and why positive? don't get it} and $d(q)$ the signed
distance between two shapes, they start with generic complementarity
$0 \le \l \quad\bot\quad d(q) \ge 0$ (equation (8) in
\cite{xie2016rigid}), where $\bot$ means that at least one of the
inequalities holds with equality. In their equations (9-13) this is
then reformulated by introducing the effective contact point as two 3D
positions $a_1$ and $a_2$, one on each shape surface, and constraining
them to be inside the convex shape polytopes with explicit
inequalities for each shape face. Analytic solutions for a single
simulation step are then provided for the surface and line contact
cases.

Our POA approach adopts the general idea of an effective contact
point, but changes the formulation to only introduce a single 3D
decision variable (namely the point of attach), formulate constraints
based on a generic signed distance function rather than an explicit
convex polytope description of shapes, and embedding the formulation
in path optimization rather than building on analytic solutions for
one step simulation.

\subsection{Task and Motion Planning (TAMP)}

Most existing TAMP approaches build on a
discretization of the configuration space or action parameter space
\cite{2014-Lozano-Perez-constraintbased,2012-Lagriffoul-Constraint,2014-Lagriffoul-Efficiently}
and/or sample-based planners
\cite{2014-Srivastava-Combined,2013-deSilva-combining,2010-Alili-Interleaving}. Our
own previous work \cite{18-toussaint-RSS} proposed Logic-Geometric Programming (LGP), an
optimization-based approach that can also plan tool-use and dynamic
interactions using simplified Newtonian equations and impulse
exchange. However, the particular physics description used in this
previous LGP formulation is not general enough to enable broader physical
reasoning. In particular, our previous work was missing force-based
contact models, proper Newton-Euler equations, and quasi-static
variants. The present paper proposes exactly such extensions and
thereby aims to consider more generally what are appropriate building
blocks in a multi-physics approach to sequential manipulation
planning. We focus on the modeling questions and our experiments
solve manipulation problems for a given skeleton.  Therefore, what is
proposed here is only a component of a complete physical TAMP solver
that also searches over skeletons, such as our Multi-Bound Tree Search
\cite{17-toussaint-ICRA}.


\subsection{Tool-use in animals and humans}

Tool-use in animals and humans was described, e.g., in
\cite{koehler:17,2009-Wimpenny-CognitiveProcessesAssociated,2016-Osiurak-Tooluseaffordance}. With
our work we aim to provide computational methods to enable such
reasoning. General physical reasoning in animals and humans is studied
and discussed in
\cite{2012-Hamrick-Physicalreasoningcomplex,2013-Vonk-SocialPhysicalReasoning,2013-Battaglia-Simulationenginephysical,2014-Seed-SpacephysicsChildren}. Particularly
interesting is the discussion of simplistic and intuitive models of
physics \cite{2018-Smith-Differentphysicalintuitions} that one might
consider as being ``incorrect'', but which are effective heuristics
for real-world reasoning and decision making. This motivated our
approach of enabling multi-physics descriptions within a coherent
planning framework.

Finally, one of our demonstration scenarios is inspired by
\cite{2018-Hogan-ReactivePlanarManipulation}, which leverages machine
learning methods to enable real-time MPC control through planar
manipulation interactions. Fig.~1 in
\cite{2018-Hogan-ReactivePlanarManipulation} describes a scenario
where a book is pulled from a shelf to enable a subsequent stable
grasp, which exploits different contact modalities and motives our
multi-physics description. We consider this scenario in section
\ref{secBook}.

\section{Trajectory Optimization Framework}

This section presents the path optimization framework as an extension
of the LGP-formulation from \cite{18-toussaint-RSS} to allow
optimizing for physical interactions.  We optimize a path $x : [0, KT]
\rightarrow \XX$ consisting of $K\in\mathbb{N}$ phases or modes.  A
discrete variable $s_k$ defines the constraints and costs in each
phase $k$ of the path.  We call a sequence of discrete variables
$s_{1:K}$ a \emph{skeleton}.  The configuration space
$\XX\subset\RRR^n\times SE(3)^m \times \RRR^{6\cdot n_\text{cp}}
\times \RRR$ includes the $n$-dimensional generalized coordinate of
robot links ($\RRR^n$) and the poses of $m$ rigid objects ($SE(3)^m$),
as in the original formulation \cite{18-toussaint-RSS}. However, in
this work a configuration additionally includes wrench interactions
for each of $n_\text{cp}$ possible contact pairs
($\RRR^{6n_\text{cp}}$) as well as a single scalar $\tau\in\RRR$ with
the following semantics: The continuous path $x$ in the configuration
space is indexed by the continuous path coordinate $t\in [0, KT]$,
which corresponds to a virtual time, not real-world time. Hence the
duration of the path in terms of the path coordinate is fixed to $KT$. To make this
consistent with physics we jointly optimize for the time scaling
$\tau$, which defines the relation between the path coordinate $t$
and real time, see Sec.~\ref{secTimeStepping} for details.

Given a skeleton $s_{1:K}$, we solve the path problem
\begin{subequations}\label{eq:LGP}
\begin{align}
\min_{x:[0,KT]\to \XX} & \int_0^{KT} f_\path(\bar x(t))~ dt  \\
\st~& x(0)=x_0,~ h_\goal(x(T))=0,~ \\
\begin{split}
& \forall_{t\in[0,T]}:~
   h_\path(\bar x(t), s_{k(t)})=0, \\
& \phantom{\forall_{t\in[0,T]}:~}
   g_\path(\bar x(t), s_{k(t)})\le0
\end{split} \\
\begin{split}
& \forall_{k\in\{1,..,K\}}:~
  h_\switch(\hat x(t_k), s_{k\1}, s_k)=0, \\
& \phantom{\forall_{k\in\{1,..,K\}}:~}
  g_\switch(\hat x(t_k), s_{k\1}, s_k)\le0,
\end{split} ~.
\end{align}
\end{subequations}
Here, $f_\path$ define path costs, which in our case are squared
accelerations of the robot joints plus additional regularizations, as
described in Section \ref{secReg}, and $h_\goal(x(T))$ define goal
equality constraints. Further, $(h,g)_\path$ define differentiable
path constraints for a given mode $s_k$ with $k(t) = \lfloor t/T
\rfloor$, which depend on $\bar x(t) = (x(t), \dot x(t), \ddot x(t))$,
and $(h,g)_\switch$ define differentiable switch constraints between
modes $s_{k\1}$ and $s_k$, which depend on $\hat x = (x, \dot x, \dot
x')$, where $\dot x$ is the velocity before, and $\dot x'$ is the
velocity after a switch (e.g.\ impulse exchange)
\cite{18-toussaint-RSS}.

To formulate physics laws in terms of real-time, the real-time
velocities and accelerations can trivially be determined by the
chain-rule, e.g.\
\begin{align}
	\frac{\diff x}{\diff \tau}(t) =  \frac{\partial x}{\partial t}(t)\frac{\diff t}{\diff \tau}(\tau(t)).
\end{align}

Our path solver, KOMO \cite{17-toussaint-Newton}, parameterizes each
configuration with only the minimal set of degrees-of-freedom
(dofs). For instance, actual optimization variables for force exchange
are only introduced when the skeleton introduces the existence of a
force interaction. Therefore, the path solver deals with varying dofs
in each mode and at each switch, which depend on the skeleton.

The following section details the specific dofs, inequality and
equality constraints that are introduced into this path optimization
formulation to by force-based interaction modes.

%

\section{Contact Models, Path Constraints, and Quasi-Static Motion}\label{secModel}

In this section we first formulate an efficient parameterization of
force interactions using the \emph{point-of-attack} (POA), and then
specific forced and complementary contact models. Finally, we give
more details on the optimizer employed.

  
\subsection{Contact interaction modes}\label{secModes}

In our approach the skeleton \cite{18-toussaint-RSS} decides between
which objects and in which phase contact interactions are accounted
for. When contact interaction between a pair of objects is created,
this has several effects on the resulting path problem: (1) A 6D
decision variable (wrench, or force and POA) for each time step is
introduced, (2) constraints are added to the path problem that
describe physically correct forces and POA in consistency with the
geometric configuration, and (3), if one of the objects are either in
quasi-static or in dynamic motion mode, the effective wrench of the
contact enters its quasi-static or dynamic motion constraint.

We provide several options to impose contacts during the
manipulation. Specifically, we allow for a forced contact (requiring
zero-distance throughout the interval), an instantaneous contact
(active at one time slice only, realizing elastic impulse exchange),
and the standard complementarity (enforcing complementarity throughout
the interval). Each of these three contact modes comes in two
versions, one allowing for slip, the other enforcing stick. We
describe the details in the following.

\subsection{Wrench as Force \& Point-of-Attack}

For each force interaction we introduce a 6D decision variable in the
path optimization problem to represent the total wrench exchange
between the two rigid bodies that enters their Newton-Euler
equation. However, instead of parameterizing a wrench $(f,\o)$ directly
as linear force $f\in\RRR^3$ and torque $\o\in\RRR^3$, we equivalently
parameterize it as $(f,p)$, where $p\in\RRR^3$ is the 3D point-of-attack
(POA) in world coordinates (or zero-momentum point), with $\o = f
\times p$. This formulation follows the idea of the equivalent contact
point \cite{xie2016rigid} and resolves several issues that arise when
deciding on force interaction during optimization.

We want to emphasize that the primary semantics of the POA is a
parameterization of the exchanged 6D wrench in the Newton-Euler
equations -- at initialization or during optimization, the POA might
well not be located on the object surfaces. Only at the point of
convergence of optimization, and only if the exchanged force is
non-zero, the POA fulfills the necessary geometric constraints
(described below) to bear the semantics of a contact point
representative. In other terms, the POA is a means to let the
optimizer try to find a wrench exchange that is eventually consistent
with both, contact geometry and the Newton-Euler equations.

This approach is in contrast to typical forward simulation models,
where the point of force exchange is assumed to be at contact points
computed from the current geometric configuration. However, contact
point(s) computed from the geometric configuration are unstable
(chaotic) for flat-on-flat interactions, cause jittering or bouncing,
and raise serious convergence issues for path optimization.

Note that using the POA does not mean we limit ourselves to only point
contacts. The POA parameterizes the wrench between two object
surfaces: by constraining it (if force is exchanged) to lie on both
object surfaces we can elegantly handle any contact configurations
(point-to-point, point-to-line, line-to-line, point-to-surface,
line-to-surface, surface-to-surface) and continuous transitions
between them, as highlighted by the example of the box sliding over an
edge. However, note that by constraining the POA to be on the surfaces
and only allowing for a linear force $f$ there, our current
implementation excludes the possibility of a torque around the normal
(e.g., from patch friction) \cite{2017-Fazeli-Parametercontactforce}
or wrenches via adhesion.

\subsection{Forced and Complementary Contacts}

We distinguish a
\emph{forced} contact and a \emph{complementary} contact. When
the skeleton imposes a forced contact, we add the constraints
\begin{align}
d_1(p) &= 0 && \text{(POA is on object 1)}\\
d_2(p) &= 0 && \text{(POA is on object 2)}\\
d_{12} &=0 && \text{(object 1 and 2 touch)} ~,
\end{align}
where $d_1(p)$ is the (signed) distance or penetration of $p$ to the
convex mesh of object 1; and $d_{12}$ is the signed distance or
penetration between two convex meshes. Both are evaluated with either
Gilbert–Johnson–Keerthi (GJK) for non-penetrating objects, and
Minkowski Portal Refinement (MPR) for penetrating objects.

When the skeleton imposes a complementary contact, we only add the 7D constraint
\begin{align}
(d_1(p)f, d_2(p)f) &=0 && \text{(force complementarity)} \\
d_{12} &\ge 0 && \text{(no penetration)} ~.
\end{align}
Note that complementarity is imposed with both(!) POA distances, not via
the numerically less stable geometric distance $d_{12}$. The
POA is directly a decision variable, with trivial Jacobian, whereas
$d_{12}$ is only piece-wise differentiable. But the combined
constraints do eventually imply complementarity w.r.t.\ object touch.

\subsection{Positivity, Slip and Elasticity}
  
We always constrain the force to be positive,
\begin{align}
  - n^\T f &\le 0 && \text{(force is positive)}
\end{align}
where $n\in\RRR^3$ is the normal of the pair's distance or penetration vector,
which we retrieve differentiably from the witness simplices.

To model stick as well as elastic bounce we impose velocity
constraints on the actual object surface points that relate to the
POA. Let $V_1 = v_1 + w_1 \times (p-p_1)$ be the object-associated POA
velocity, where $(v_1,w_1)$ are the linear and angular velocities of
object 1, and $p_1$ its center. Let $V = V_1 - V_2$ be the relative
POA velocity between the interacting objects 1 and 2.
For a non-slip contact we impose the equality constraint
\begin{align}
  (\Id-nn^\T) V &=0 && \text{(zero tangential surface velocities)}
\end{align}
and the inequality constraint
\begin{align}
||(\Id-nn^\T) f||^2 &< \mu^2||n^\T f||^2 && \text{(quadratic friction cone)}
\end{align}	
where $\mu$ is the coefficient of friction in Coulomb's friction model.
In contrast, for sliding contacts, the force $f$ must be on the edge of the friction cone
and its tangential component needs to align with the negative relative POA velocity, which both is ensured by
\begin{align}
(\Id-nn^\T) f &= -a(\Id-nn^\T) V, &&
\end{align}
for $a = \mu |n^\T f|/|V|$. Note that in our experiments we only consider very large (stick) or low (slip) friction. For friction-less contact this implies a normal force
\begin{align}
  (\Id-nn^\T) f &= 0 && \text{(force is normal)}.
\end{align}

Finally, let $V'$ be the relative POA velocity one time step later --
e.g., after an instantaneous bounce. For an instantaneous bounce with
elasticity coefficient $\b$ we add the constraint (cf.~\cite{stewart2000implicit})
\begin{align}
  n^\T (V + \b V') &=0 && \text{(normal velocity reflection)}
\end{align}

In summary, using these equations we can impose forced, instantaneous,
and complementary contacts, each with slip or stick.

\subsection{Regularization costs}\label{secReg}

While we impose hard constraints to ensure physical correctness, we
additionally can have soft penalties to favor smooth
interactions. This can be interpreted as a prior on which kinds of robot
manipulations we favor -- for instance, those where the POA does not
exceedingly jump around. Adding such regularizations has a 
positive effect on the convergence behavior of the
solver. Specifically we add cost terms
\begin{align}
  \norm{\ddot p}^2 &&& \text{ (POA acceleration penalization)} \\
  \norm{\ddot f}^2 &&& \text{ (force acceleration penalization)} \\
  \norm{f}^2 &&& \text{ (force penalization)} \\
  \norm{V_1 - V_2}^2 &&& \text{ (sliding velocity penalization)} ~.
\end{align}

\subsection{Dynamic and Quasi-Static Motion}

For every object that is in dynamic or quasi-static motion mode we
collect the total 6D wrench $F$ on its center that arises from all contacts
with their forces and POAs. From the current path we compute the object's
linear and angular velocity $(v,w)$ and acceleration $(\dot v, \dot
w)$. Given the gravity vector $g\in\RRR^6$, the inertia matrix
$M\in\RRR^{6\times 6}$ and a potential friction $\l$, we
have the Newton-Euler equation,
\begin{align}\label{eqNE}
  \mat{c}{\dot v\\\dot w} + g - M^\1 (F - \l) &=0 && \text{(Newton-Euler)}.
\end{align}
For free flight objects we assume $\l=0$. 

For the quasi-static sliding, we assume some friction $\l$ such that the inertia forces can be ignored $(\dot v, \dot w) \equiv 0$.
In particular, when an object is pushed by a manipulator on a table, this quasi-static model constrains the object's motion to be only along the surface of the table, $(v_z, w_{\phi,\psi}) = 0$. In the remaining degrees of freedom, the wrench applied by the manipulator exactly cancels out the friction, i.e., $F'= F_{x,y,\theta} = \l_{x,y,\theta}$, and is related to the velocity as
\begin{align}
  \mat{c}{v_{x,y}\\w_{\theta}} &= k\nabla H(F')
\end{align}
with a convex function $H(F'):\RRR^3\to[0,\infty]$.
Note that, in this case, we need to consider the scaled wrench $kF'$ as optimization variables instead of $F'$~\cite{zhou2018convex,halm2018quasi}.
In the experiment, we assumed the pressure is distributed constantly, thus utilized a simple quadratic representation of $H$: $H(F') = \frac{1}{2}(F')^T(M')^\1F'$.

\subsection{Time scale optimization and impulse vs.\ force exchange}\label{secTimeStepping}

Finally, for the case of truly dynamic sequences, e.g.\ where a ball
is bouncing on a table, it is physics that decides on the true time
between two interactions. However, the skeleton imposes
interactions at fixed steps. To resolve this conflict we introduce
a scalar decision variable $\tau_t \in\RRR$ for every discretization
step of the path which represents the real-time step in seconds. The
optimizer can thereby find $\tau_t$ that scales a fixed path section
to a correct physical time interval. We impose positive time
evolution, $\tau_t \ge 0$, choose piece-wise constant time scaling
$\dot\tau_t=0$ within phases, and introduce a regularization
$\norm{\tau_t - \hat \tau}^2$ to favor time steps close to the
default. All velocities mentioned above are differentiably evaluated
by finite differencing along the path and dividing by $\tau_t$. The
acceleration terms $(\dot v,\dot w) + g$ in the Newton-Euler equation
are actually multiplied by $\tau_t$, which semantically makes it an
impulse exchange equation, and the decision variable $f$ associated
with contacts actually represent impulses.

\subsection{Integration with Existing Solver and Stable Interaction Models}

The models described above were integrated in the existing solver
described in \cite{18-toussaint-RSS}. This means that the above
interaction modes can be mixed with modes for stable grasping and
placement of objects. We also adopt the switch constraints, which
enforce zero object accelerations at the switch, except for
instantaneous contacts. As stated previously, in this work we focus on
the path problem for a given skeleton, which is represented as a list
of first order literals.

The resulting optimization problem is a non-linear mathematical
program, which we address with an Augmented Lagrangian method
\cite{17-toussaint-Newton} that exploits the sparse structure of the
global path Hessian when computing Newton steps. We initialize the
solver with the constant path (no object is moving) plus small
Gaussian noise (0.01 sdv in joint space) to break symmetry. Restarts
are used to find good local optima. Additional decision variables
that are added due to the interactions (see Sec.~\ref{secModes}) are
initialized to match the initial scene poses (e.g., grasps have huge
offset between endeffector and object). The introduced wrench decision
variables are initialized as follows: linear forces are initialized as
zero, while the POA is initialized as mean between the centers of the
interacting objects.



\section{Experiments}\label{secExperiments}

Please see the accompanying
video\footnote{\url{https://youtu.be/tVFkKIIODaM}} to get a first
impression on our results.  The source code to reproduce all examples
in the video is
available\footnote{\url{https://github.com/MarcToussaint/20-IROS-ForceBased}}. All
experiments were done on an Intel i7-6500U CPU @ 2.50GHz.

\subsection{Passive Tests (Pure Simulation as Path Optimization)}

We start with first reporting on tests where there exists no robot or
actuators, but path optimization is merely used to compute a
physically feasible path. This tests whether our descriptions of
physical interactions are appropriate to also perform ordinary
physical simulation using path optimization.\footnote{In fact, a
  physical simulator could be implemented using MPC based on our
  formulation, repeating path optimization of a short receding
  horizon. This would enable features such as co-optimizing the time
  stepping $\tau$ for the sake of simulation precision, or
  equality-constraining the simulation additionally on precise
  long-term energy conservation. We haven't explored further in this
  direction.}  While the tests are trivial scenarios, they give
interesting insights in the method.

\subsubsection{Ball bouncing}

\begin{figure}\centering
  \begin{subfigure}[b]{0.3\columnwidth}\centering
  \showh[1]{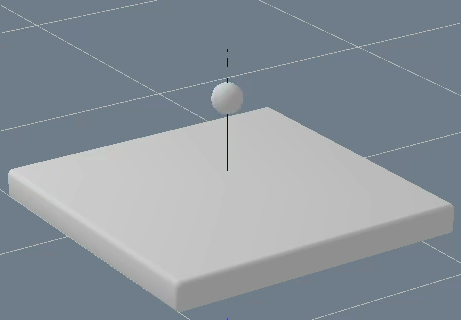}
  \caption{}
  \end{subfigure}
  \begin{subfigure}[b]{0.3\columnwidth}\centering
    \showh[1]{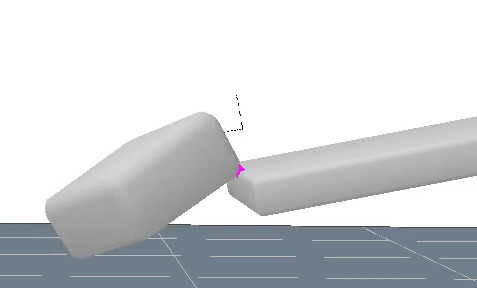}
    \caption{}
  \end{subfigure}
  \begin{subfigure}[b]{0.3\columnwidth}\centering
    \showh[1]{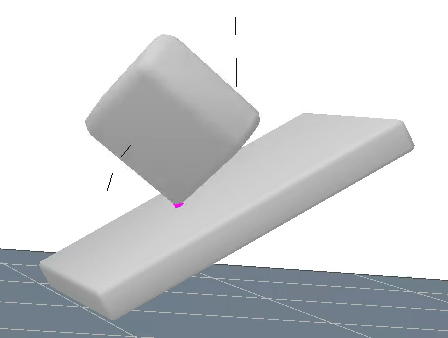}
        \caption{}
  \end{subfigure}

  \begin{subfigure}[b]{0.7\columnwidth}\centering
  \hspace*{-7mm}\showh[1]{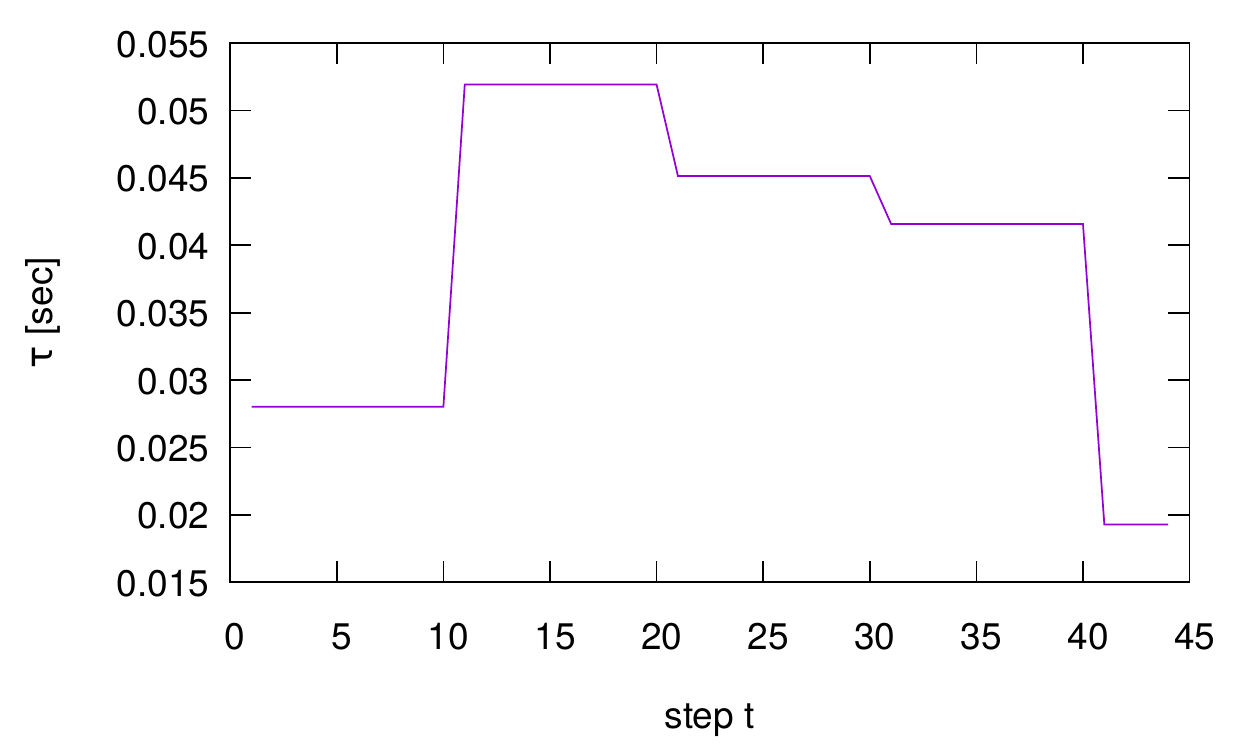}
  \caption{}
  \end{subfigure}

  \caption{\label{figTau} Three passive simulation tests. (a) Bounding
    ball, (b) sliding over the edge, (c) tumbling box. (d) shows
    the time stepping optimization for (a).}
\end{figure}

We drop a ball onto a table (Fig.~\ref{figTau}(a)), letting it bounce
4 times with elasticity coefficient $\b=0.9$ (where the outgoing
velocity is constrained to be 90\% of the incoming velocity). The
accompanying video shows the simple behavior. The optimizer robustly
converges within about $0.26s$ to standard precision, and within about
$1$ second to ultimate floating point precision in the constraints.

An insight we gain from this is that, in order to compute correct physical bounces, it is essential to include co-optimization of the time stepping $\tau$. Fig.~\ref{figTau}(d) shows $\tau_t$ for time steps $t=1,..,T$, for $T=45$. We see that the optimizer found different time scalings during each bounce interval. This is essential as the duration of the bounce is determined by physics and must be aligned with the imposed bounce schedule. This also means that those configurations at which bouncing contact is imposed are optimized to be at exactly the real times where the ball hits the table, allowing all constraints to be fulfilled to arbitrary precision even though we choose a very coarse time discretization. This addresses the typical issue in physical simulations of choosing efficient but imprecise fixed time steps versus adaptive step-sizes. Disabling stepping optimization makes our approach fail to find a correct solution for this simple bouncing problem.

\subsubsection{Slide-falling and tumbling block}

We present two passive examples that highlight the POA mechanism, a
box sliding from a tilted table (solver time $2.41s$,
Fig.~\ref{figTau}(b)) and a box tumbling with sticky contact on a
tilted table (solver time $0.49s$, Fig.~\ref{figTau}(c)). In both
cases we used general complementary contacts, and it was essential to
allow the optimizer to find a suitable POA. Without the POA decision
variable (when inserting the central witness point of the current
configuration between the current shapes, computed with GJK or MPR,
instead of $p$ in all constraints), the solver was unable to find
solutions in both scenarios. Using the POA, the optimizer finds
(mostly smooth) POA movements on the surfaces.

\subsection{Physical Manipulations}

\begin{figure*}\centering
  \begin{subfigure}[b]{0.19\textwidth}\centering
  \showh[1]{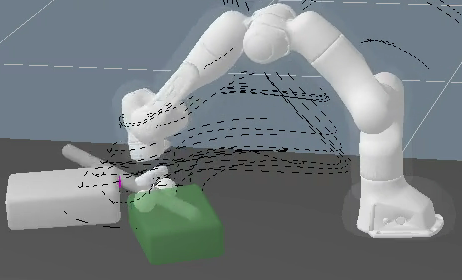}
  \caption{}
  \end{subfigure}
  \begin{subfigure}[b]{0.19\textwidth}\centering
    \vspace*{3mm}
    \showh[1]{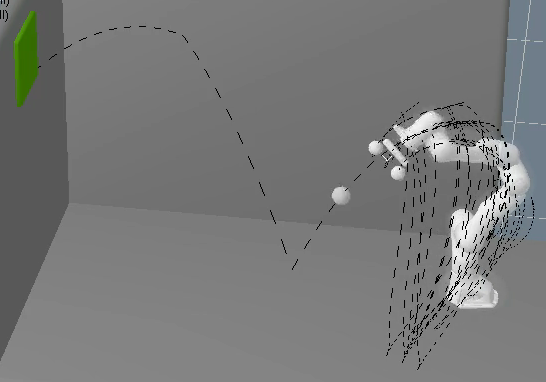}
        \caption{}
  \end{subfigure}
  \begin{subfigure}[b]{0.19\textwidth}\centering
    \showh[1]{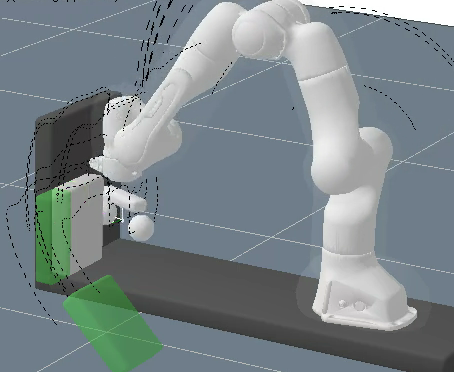}
    \caption{}
  \end{subfigure}
  \begin{subfigure}[b]{0.19\textwidth}\centering
    \showh[1]{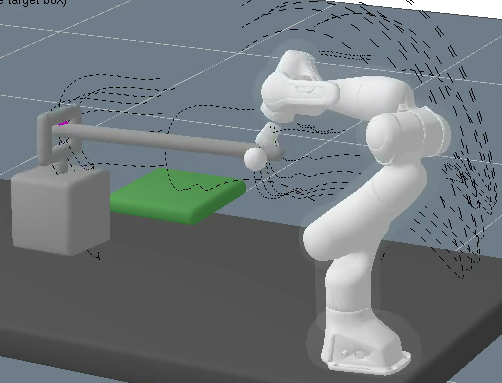}
        \caption{}
  \end{subfigure}
  \begin{subfigure}[b]{0.19\textwidth}\centering
    \vspace*{2mm}
    \showh[.9]{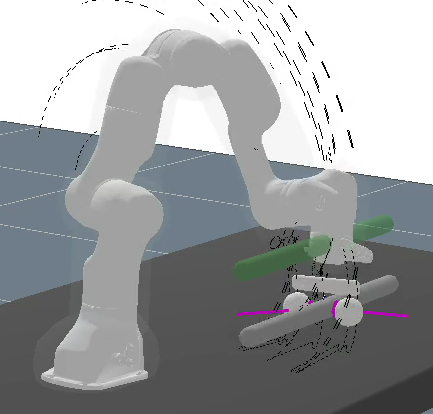}
        \caption{}
  \end{subfigure}

  \caption{\label{figRob}
  Physical robot manipulation demonstrations: (a) Picking a stick to push-rotate a box to a target pose; (b) picking a ball to throw it onto the floor so that it bounces back against a wall, and then bounces to a given target; (c) pushing a book forward to become graspable, pick it and move it to a target; (d) picking a stick to lift a box through a ring in order to place it onto a target; (e) basic force-based lifting a stick.}
\end{figure*}

In the remainder we consider sequential robot manipulation
scenarios. In all cases the manipulator model is a Franka Emika
Panda. However, we abstracted the gripper's fingers as spheres. In
the context of stable grasping, this is motivated by our experience
that modeling the actual gripper closing with the actual finger
geometries is hardly indicative of grasp success in real-world
execution. Instead, ensuring a central opposing positioning to normal
surfaces is simpler and transfers well. Abstracting the two fingers as
spheres allows us to constrain that the nearest distance vector from
the left finger-sphere to the object should exactly oppose the nearest
distance vector from the right finger-sphere to the object. To this
end we constrain the sum of both vectors to be zero, which describes a
central opposition and has nice gradients to pull the gripper towards
an opposing grasp.


\subsubsection{Quasi-static pushing with a picked stick}

In this scenario (Fig.~\ref{figRob}(a)) the robot picks up a stick in order to push the box to the green target pose. The box motion is modeled as quasi-static table sliding. The pre-defined skeleton is
\begin{code}
\begin{verbatim}
(oppose finger1 finger2 stick) (stable gripper stick)
(quasiStaticOn table box) (contact_slide stick box)
(poseEq box target)
\end{verbatim}
\end{code}%
where each line corresponds to one phase step, the predicates
\pre{stable} and \pre{quasiStaticOn} describe our mode switches,
\pre{contact\_slide} the creation of a forced sliding contact, and
\pre{oppose} and \pre{poseEq} are geometric constraints.

The solver finds (in $31.47s$) a rather involved pushing maneuver where the POA between the stick and the box is controlled to places that allow pushing the box into different directions. The video displays several additional pushing sequences, some with a free floating gripper, to show the variety of solutions found by the solvers. This scenario and the following two are cases where the solver benefits from mixing physics descriptions of varying abstraction: the stable grasp abstraction for the interaction with the stick, and force-based modeling for the interaction between box and stick, and quasi-static dynamics for the box. Our last experiment will investigate the gained efficiency of a stable grasp abstraction vs.\ a force-based grasp.

\subsubsection{Dynamic ball throwing and bouncing to a target}

This scenario (Fig.~\ref{figRob}(b)) is an extension of the passive bouncing test discussed above. A robot picks up a ball to throw it onto the floor so that it bounces back against a wall, and then bounces to a given target. This highlights the ability to implicitly propagate back target constraints through force-based contacts to yield a correct throwing strategy. The pre-defined skeleton is
\begin{code}
\begin{verbatim}
(oppose finger1 finger2 ball) (stable gripper ball)
(dynamic ball)
(bounce ball table)
(bounce ball wall)
(touch target ball)
\end{verbatim}
\end{code}%
which states that the first phase ends with grasping the ball, the ball becomes free and dynamic (Newton-Euler equations) after the second phase, the ball bounces with the table after the third, with the wall after the fourth, and touches (zero distance) the green target after the fifth.

The solver finds a solution (in $16.3s$) where the robot, after picking up the ball, nicely accelerates and releases the ball to bounce to the target, as in 3D billiards. The velocity of the full sequence has to be rather fast as the free ball flight is governed by physics. Therefore, the control costs of this path are highly significant in this optimization problem. As seen in the video, the found solution varies drastically depending on the scaling of control costs.

\subsubsection{Using a stick to lifting a weight at a ring to place it on a target}

This scenario (Fig.~\ref{figRob}(c)) aims to highlight the ability to create the needed
contact points to achieve long term targets. A robot grasps a stick in
order to insert it into a ring at the top of a weight. Thereby it can
lift it and transport it onto a given target. The given skeleton is
\begin{code}
\begin{verbatim}
(oppose finger1 finger2 stick) (stable gripper stick)
(dynamic box) (contact_slide stick ring)
(stableOn target box) (above target box)
\end{verbatim}
\end{code}%
where \pre{dynamic} switches the box to free flying mode (Newton-Euler
equations with contact forces as input), while \pre{stableOn} then
switches the box to reside stably on the target. \pre{above}
geometrically constraints the box center of mass to be within the
target support.

The solver finds (in $10.0s$) a solution, where the robot finds the
right spot for the stick to touch the ring so as to lift it. The box
swings slightly during the dynamic transport to the target. The
transition from its initial resting on the table to the dynamic phase
is not perfectly smooth, which would require more careful
regularization of accelerations at mode switches.


\subsubsection{First pushing then grasping a book from a shelf}\label{secBook}

This scenario (Fig.~\ref{figRob}(d)) is inspired from Fig.~1 in \cite{2018-Hogan-ReactivePlanarManipulation} and considers a book on a shelf that is initially too close to a wall to be grasped. So it first has to be pushed forward in order to allow to grasp it. The pre-defined skeleton is
\begin{code}
\begin{verbatim}
(contact finger1 book) (quasiStaticOn shelf book)
(poseEq book subTarget)
(oppose finger1 finger2 book) (stable gripper book)
(poseEq book target)
\end{verbatim}
\end{code}%
Note that in this skeleton we predefined an intermediate target pose
for the book, the first green pose seen in the video, as discussed in
detail below.

Given this skeleton, the solver finds (in $16.4s$) a solution where
the robot places the finger nicely to the right of the book to push it
to the sub-target, then in a minimal motion transitions to the
opposing grasp to lift the book and carry it to the final target.

As a negative result, if we remove the sub-target (2nd line) from the
skeleton, the solver fails to find a feasible solution and typically
converges to an infeasible solution that cheats when picking up the
object, squeezing the finger between wall and book in a penetrating
and book-jumping manner (see video). We considered extensively how we
could fix this deficit of our method. However, we concluded that
without cheating by redesigning the scenario to become less symmetric
and have an intrinsic bias towards the first book slide, there is no
way for our approach to solve this problem without introducing the
sub-goal or some similar bias. The path optimization process has no
implicit gradient towards paths that have consistent book
slides in one or another direction. A random initialization is too
unsystematic to pit optimization towards such slides. Instead, due to
symmetry, path optimization is most likely to converge to the local
optimum that corresponds to the shown infeasible solution.

We believe this scenario is highly insightful. Local optima are a
fundamental issue for optimization and source of complexity for
planning. The scenario shows that stronger biases would have to
pre-exist, perhaps have been learned, to solve complex manipulation
problems.

\subsubsection{Force-based vs.\ stable grasping}

In the previous experiments we imposed stable grasps. We can also
solve for force-based grasping. In the last scenario
(Fig.~\ref{figRob}(e)) the robot only needs to lift the stick to a
target pose. For the pre-defined skeleton with force-based contacts
\begin{code}
\begin{verbatim}
(oppose fing1 fing2 stick) (contactStick fing1 stick)
  .. (contactStick finger2 stick) (dynamic stick)
(poseEq stick target)
\end{verbatim}
\end{code}%
it takes the solver $5.4s$ to find a lift. However, for the skeleton
with stable grasp
\begin{code}
\begin{verbatim}
(oppose finger1 finger2 stick) (stable gripper stick)
(poseEq stick target)
\end{verbatim}
\end{code}%
the solver requires only $0.24s$.






\section{Conclusions}

In this work we propose concrete models for physical reasoning and
robot manipulation planning which allow the solver to mix different
abstractions for different objects and phases of the solution, and
integrate this in a path optimization framework to solve sequential
physical manipulation problems over a wide range of scenarios. We call
this a \emph{multi-physics} model for reasoning. Our solver is based
on a path description of physics that directly allows us to
leverage constrained optimization methods.

A limitation of the approach is the still significant computation time
needed to solve complex sequential physical interaction scenarios
($10-40s$ in our examples). This makes the naive integration into the
full symbolic search of LGP unattractive. A promising alternative is
to use our solver to generate large-scale data to learn a heuristic
that can drastically accelerate search over potential interaction
skeletons \cite{driess20deep}. Further, this work only considers the
problem of reasoning about possible manipulation sequences, not
controller synthesis for a robust execution of such plans. Translating
the framework to stochastic optimal control is yet subject to research
\cite{ha20probabilistic}.



\section*{Acknowledgments}

We would like to thank Alberto Rodriguez for inspiring us to work on
some of the problems. This work was funded by the Baden-W\"urttemberg
Stiftung in the scope of the NEUROROBOTICS project
\emph{DeepControl}. M.T.\ thanks for the Max Planck Fellowship at the
MPI for Intelligent Systems.


\bibliographystyle{IEEEtran}

\bibliography{19-forceBased,19-physicalReasoning-humans,17-LogicGeoProgramming,17-optimization,group,refs}

\end{document}

%% file: macros.tex
  \renewcommand{\b}{\beta}
  \renewcommand{\d}{\delta}

    \newcommand{\g}{\gamma}
    
  \renewcommand{\l}{\lambda}
  \renewcommand{\L}{\Lambda}
    \newcommand{\m}{\mu}

  \renewcommand{\k}{\kappa}
  \renewcommand{\o}{\omega}
  \renewcommand{\O}{\Omega}

  \renewcommand{\r}{\varrho}
    \newcommand{\s}{\sigma}
  
  \renewcommand{\t}{\theta}
    \newcommand{\T}{\Theta}

  \renewcommand{\SS}{{\cal S}}

    \newcommand{\XX}{{\cal X}}

  \newcommand{\RRR}{{\mathbb{R}}}

  \renewcommand{\|}{\,|\,}
  
  \renewcommand{\=}{\!=\!}

  \newcommand{\Id}{{\rm\bf I}}

  \newcommand{\w}{\wedge}

  \newcommand{\st}{\quad\text{s.t.}\quad}

  \newcommand{\mat}[3][.9]{
    \renewcommand{\arraystretch}{#1}{\scriptscriptstyle{\left(
      \hspace*{-1ex}\begin{array}{#2}#3\end{array}\hspace*{-1ex}
    \right)}}\renewcommand{\arraystretch}{1.2}
  }

  \providecommand{\href}[2]{{\color{blue}USE PDFLATEX!}}
  \providecommand{\url}[2]{\href{#1}{{\color{blue}#2}}}
  \newcommand{\anchor}[2]{\begin{picture}(0,0)\put(#1){#2}\end{picture}}

  \newcommand{\hide}[1]{
    \begin{list}{}{\leftmargin0ex \rightmargin0ex \topsep0ex \parsep0ex}
       \helvetica{5}{1}{m}{n}
       \renewcommand{\section}{\par SECTION: }
       \renewcommand{\subsection}{\par SUBSECTION: }
       \item[$~~\blacktriangleright$]
       #1
       \message{^^JHIDE--Warning!^^J}
    \end{list}
  }

  \newcommand{\fullhide}[1]{\message{^^JHIDE--Warning (hidden)!^^J}}

  \newcommand{\mytitle}[1]{\title{#1}\newcommand{\thetitle}{#1}}
  \newcommand{\header}{\begin{document}\mytitle\cleardefs}
  \newcommand{\contents}{{\tableofcontents}\renewcommand{\contents}{}}
  \newcommand{\footer}{\small\bibliography{marc,bibs}\end{document}}
  \newcommand{\widepaper}{\usepackage{geometry}\geometry{a4paper,hdivide={25mm,*,25mm},vdivide={25mm,*,25mm}}}
  \newcommand{\moviex}[2]{\movie[externalviewer]{#1}{#2}} 
  \newcommand{\rbox}[1]{\fboxrule2mm\fcolorbox[rgb]{1,.85,.85}{1,.85,.85}{#1}}
  \newcommand{\mpage}[2]{{\begin{minipage}{#1\columnwidth}#2\end{minipage}}}
  \newcommand{\redbox}[2]{\fboxrule1mm\fcolorbox[rgb]{1,.7,.7}{1,.7,.7}{\begin{minipage}{#1\columnwidth}\center#2\end{minipage}}}
  \newcommand{\onecol}[2]{
    \begin{minipage}[c]{#1\columnwidth}#2\end{minipage}}
  \newcommand{\twocol}[5][0]{
    \begin{minipage}[c]{#2\columnwidth}#4\end{minipage}\hspace*{#1\columnwidth}%
    \begin{minipage}[c]{#3\columnwidth}#5\end{minipage}}
  \newcommand{\threecol}[7][0]{%
    \begin{minipage}[c]{#2\columnwidth}#5\end{minipage}\hspace*{#1\columnwidth}%
    \begin{minipage}[c]{#3\columnwidth}#6\end{minipage}\hspace*{#1\columnwidth}%
    \begin{minipage}[c]{#4\columnwidth}#7\end{minipage}}
  \newcommand{\threecoltext}[7][c]{
    \begin{minipage}[#1]{#2\textwidth}#5\end{minipage}%
    \begin{minipage}[#1]{#3\textwidth}#6\end{minipage}%
    \begin{minipage}[#1]{#4\textwidth}#7\end{minipage}}
  \newcommand{\threecoltop}[7][0]{%
   \begin{minipage}[t]{#2\columnwidth}#5\end{minipage}\hspace*{#1\columnwidth}%
   \begin{minipage}[t]{#3\columnwidth}#6\end{minipage}\hspace*{#1\columnwidth}%
   \begin{minipage}[t]{#4\columnwidth}#7\end{minipage}}
  \newcommand{\fourcol}[9][0]{%
   \begin{minipage}[c]{#2\columnwidth}#6\end{minipage}\hspace*{#1\columnwidth}%
   \begin{minipage}[c]{#3\columnwidth}#7\end{minipage}\hspace*{#1\columnwidth}%
   \begin{minipage}[c]{#4\columnwidth}#8\end{minipage}\hspace*{#1\columnwidth}%
   \begin{minipage}[c]{#5\columnwidth}#9\end{minipage}}
  \newcommand{\helvetica}[4]{\setlength{\unitlength}{1pt}\fontsize{#1}{#1}\linespread{#2}\usefont{OT1}{phv}{#3}{#4}}
  \newcommand{\helve}[1]{\helvetica{#1}{1.5}{m}{n}}
  \newcommand{\german}{\usepackage[german]{babel}\usepackage[utf8]{inputenc}}
\newcommand{\sans}{\renewcommand{\familydefault}{\sfdefault}}

\newcommand{\norm}[1]{|\!|#1|\!|}
\newcommand{\expr}[1]{[\hspace{-.2ex}[#1]\hspace{-.2ex}]}

\newcommand{\Jwi}{J^\sharp_W}
\newcommand{\THi}{T^\sharp_H}
\newcommand{\Jci}{J^\natural_C}
\newcommand{\hJi}{{\bar J}^\sharp}
\renewcommand{\|}{\,|\,}
\renewcommand{\=}{\!=\!}
\newcommand{\myminus}{{\hspace*{-.0pt}\text{\rm -}\hspace*{-.5pt}}}
\newcommand{\myplus}{{\hspace*{-.0pt}\text{\rm +}\hspace*{-.5pt}}}
\newcommand{\1}{{\myminus1}}
\newcommand{\2}{{\myminus2}}
\newcommand{\3}{{\myminus3}}
\newcommand{\mT}{{\text{\rm -}\hspace*{-1pt}\top}}
\newcommand{\po}{{\myplus1}}
\newcommand{\pt}{{\myplus2}}
\renewcommand{\T}{{\!\top\!}}
\newcommand{\xT}{{\underline x}}
\newcommand{\uT}{{\underline u}}
\newcommand{\zT}{{\underline z}}
\newcommand{\Sum}{\textstyle\sum}
\newcommand{\Int}{\textstyle\int}
\newcommand{\Prod}{\textstyle\prod}

\newenvironment{centy}{
\vspace{15mm}
\large
\hspace*{5mm}
\begin{minipage}{8cm}\it\color{blue}
}{
\end{minipage}
}

\newcommand{\old}{{\text{old}}}
\newcommand{\new}{{\text{new}}}
\newcommand{\MAP}{{\text{MAP}}}
\newcommand{\ML}{{\text{ML}}}

\newcommand{\redArrow}{\quad\anchor{0,-1}{\includegraphics[scale=.5]{figs/redArrow}}}
\newcommand{\pub}[1]{{\color{green}\helvetica{8}{1.3}{m}{n}#1\\}}
\DeclareMathOperator{\opKL}{KL}
\newcommand{\KL}[2]{\opKL\big(#1\,\big|\!\big|\,#2\big)} 

\renewcommand{\show}[2][.8]{\centerline{\includegraphics[width=#1\columnwidth]{#2}}}
\newcommand{\showh}[2][.8]{\includegraphics[width=#1\columnwidth]{#2}}
\newcommand{\shows}[2][.8]{\centerline{\includegraphics[scale=#1]{#2}}}
\newcommand{\showhs}[2][.8]{\includegraphics[scale=#1]{#2}}
\newcommand{\mov}[2]{\movie[externalviewer]{{\color{blue}\small #1}}{movies/#2}}
\newcommand{\movex}[2]{\movie[externalviewer]{#1}{#2}} 
\newcommand{\movh}[3][autostart,loop]{
\movie[#1]{\showh[#2]{#3.jpg}}{#3.mp4}%
\movie[externalviewer]{$\circ$}{#3.mp4}
}
\newcommand{\movc}[3][autostart,loop]{\centerline{\movh[#1]{#2}{#3}}}
\newcommand{\mymovie}[3][]{\centerline{\movie[autostart,loop,#1]{\includegraphics[#1]{#2}}{#3}}}

\newcommand{\cen}[1]{\centerline{#1}}

\newcommand{\citing}[1]{
{\color{citcol}\ttiny#1\par}
}

\newcommand{\cit}[3]{
\par\smallskip
{\color{citcol}\ttiny #1: \emph{#2}. #3 \par}
}

\newcommand{\citurl}[4]{
\par\smallskip
{\color{citcol}\ttiny #1: \protect{\href{#4}{\color{blue}{#2.}}} #3 \par}
}

\newcommand{\cito}[3]{
\par\smallskip
{\color{bluecol}\ttiny #1: \emph{#2}. #3 \par}
}

\newcommand{\redoMacrosInProof}{
  \renewcommand{\d}{\delta}
  \renewcommand{\=}{\!=\!}
}

\newcommand{\pdflatex}{
  \definecolor{bluecol}{rgb}{0,0,.5}
  \usepackage[
    colorlinks,
    urlcolor=bluecol,
    citecolor=black,
    linkcolor=bluecol,
    pdfborder={0 0 0},
    pdfpagemode=UseOutlines, 
    pdfauthor={Marc Toussaint}
  ]{hyperref}
  \DeclareGraphicsExtensions{.pdf,.png,.jpg,.eps}
  \renewcommand{\r}{\varrho}
  \renewcommand{\l}{\lambda}
  \renewcommand{\L}{\Lambda}
  \renewcommand{\s}{\sigma}
  \renewcommand{\b}{\beta}
  \renewcommand{\d}{\delta}
  \renewcommand{\k}{\kappa}
  \renewcommand{\t}{\theta}
  \renewcommand{\O}{\Omega}
  \renewcommand{\o}{\omega}
  \renewcommand{\SS}{{\cal S}}
  \renewcommand{\=}{\!=\!}
}

\newcommand{\codefont} {\helvetica{8}{1.2}{m}{n}}
\newenvironment{code}{
  \par\vspace*{-1ex}
  \codefont
}{
  \par\vspace{-1ex}\noindent%
}

\newcommand{\refeq}[1]{(\ref{#1})}

\newcommand{\signature}{
  \includegraphics[height=6ex]{signatureColor} 
}

\usepackage{algorithm}
\usepackage{algpseudocode}
\algrenewcommand{\algorithmicrequire}{\textbf{Input:~~}}
\algrenewcommand{\algorithmicensure}{\textbf{Output:}}
\algrenewcommand{\algorithmiccomment}[1]{\qquad\hfill~\hspace*{-5ex}\textit{// #1}}
\algrenewcommand{\alglinenumber}[1]{\helvetica{6}{1.3}{m}{n}#1:}

\newenvironment{algo}[1][8]{
\quad\begin{minipage}{.8\columnwidth}\helvetica{#1}{1.3}{m}{n}
\medskip\hrule\medskip
\begin{algorithmic}[1]
}{
\end{algorithmic}
\medskip\hrule\medskip
\end{minipage}
}

\newcommand{\myverbatim}{
  \usepackage{fancyvrb}
  \DefineShortVerb{\@}
  \fvset{numbers=left,xleftmargin=5ex}
}

\newenvironment{itemS}[1][4ex]{
\par
\tiny
\begin{list}{--}{\leftmargin#1 \rightmargin0ex \labelsep1ex
  \labelwidth2ex \topsep0pt \parsep0ex \itemsep0pt}
}{
\end{list}
}

\newenvironment{items}{
\par
\small
\begin{list}{--}{\leftmargin4ex \rightmargin0ex \labelsep1ex \labelwidth2ex
\topsep0pt \parsep0ex \itemsep3pt}
}{
\end{list}
}

\newenvironment{iitems}{
\par
\small
\begin{list}{--}{\leftmargin5ex \rightmargin0ex \labelsep1ex \labelwidth2ex
\topsep0pt \parsep0ex \itemsep3pt}
}{
\end{list}
}

